\documentclass{article}
\usepackage{spconf,amsmath,graphicx}
\usepackage{multirow,amssymb}
\usepackage{flushend}

\usepackage[capitalise]{cleveref}

\usepackage{subcaption}

\usepackage{enumitem}
\setlist{nosep, leftmargin=14pt}

\usepackage{mwe} 

\usepackage{float} 

\title{A Knowledge Distillation framework for Multi-Organ Segmentation of Medaka Fish in Tomographic Image}
\name{Jwalin Bhatt$^{1*}$\thanks{J. Bhatt and Y. Zharov - These authors contributed equally to this work.}, Yaroslav Zharov$^{2,3*}$, Sungho Suh$^{1,4}$\sthanks{Corresponding author: sungho.suh@dfki.de}, Tilo Baumbach$^{2,5}$, Vincent Heuveline$^{3}$, Paul Lukowicz$^{1,4}$}
\address{$^{1}$ Department of Computer Science, RPTU Kaiserslautern-Landau, Germany\\
$^{2}$ LAS, Karlsruhe Institute of Technology, Germany; $^{5}$ IPS, Karlsruhe Institute of Technology, Germany\\
$^{3}$ Engineering Mathematics and Computing Lab, Heidelberg University, Germany\\
$^{4}$ German Research Center for Artificial Intelligence (DFKI), Kaiserslautern, Germany}
\begin{document}
%
\maketitle
\begin{abstract}
Morphological atlases are an important tool in organismal studies, and modern high-throughput Computed Tomography (CT) facilities can produce hundreds of full-body high-resolution volumetric images of organisms. However, creating an atlas from these volumes requires accurate organ segmentation. In the last decade, machine learning approaches have achieved incredible results in image segmentation tasks, but they require large amounts of annotated data for training. In this paper, we propose a self-training framework for multi-organ segmentation in tomographic images of Medaka fish. We utilize the pseudo-labeled data from a pretrained Teacher model and adopt a Quality Classifier to refine the pseudo-labeled data. Then, we introduce a pixel-wise knowledge distillation method to prevent overfitting to the pseudo-labeled data and improve the segmentation performance. The experimental results demonstrate that our method improves mean Intersection over Union (IoU) by 5.9\% on the full dataset and enables keeping the quality while using three times less markup.
\end{abstract}
\section{Introduction}
\label{sec:intro}

Studies of model organisms were initially carried out through dissection or visual observation of transparent organisms such as \emph{Zebrafish}.
In recent years, the growing capacity to produce X-Ray Computed Tomography (CT) led to a qualitative change in the available data. 
With synchrotron-based micro-CT, it is not possible to produce images with $\mu$m-scale pixel size, allowing for the serial acquisition of tens of volumes without human interaction \cite{sombke2015potential, Weinhardt2018} 

Fishes, especially \emph{Medaka (Oryzias Latipes)}, have become an indispensable model organism for studying gene function in vertebrates.
The digital morphological atlas of the adult \emph{Medaka} fish allows biologists to analyze the morphometric properties of internal and external features, including organs and tissues.
The quantitative description of the organ positions and shapes can be discovered by solving the segmentation task.

The conventional technique used to generate the 3D anatomical atlas for the Medaka fish was using the semi-automated segmentation software with the help of the readily available annotations and atlases \cite{kinoshita2009medaka, shanthanagouda2014japanese, anken1998brain, bryson2007fishnet}.
This method, however, required an immense amount of hand work even for small data sets and lacks scaling abilities.
To handle these limitations, an atlas-based approach \cite{Weinhardt2018} was proposed to automate the segmentation of new samples, but it may suffer from quality loss when there are significant morphological differences between the new sample and the base segmentation.
Deep Learning can greatly help solve this problem, however, Neural Networks require extensive datasets for training.

\begin{figure*}[!t]
    \centering  
        \includegraphics[width=\linewidth]{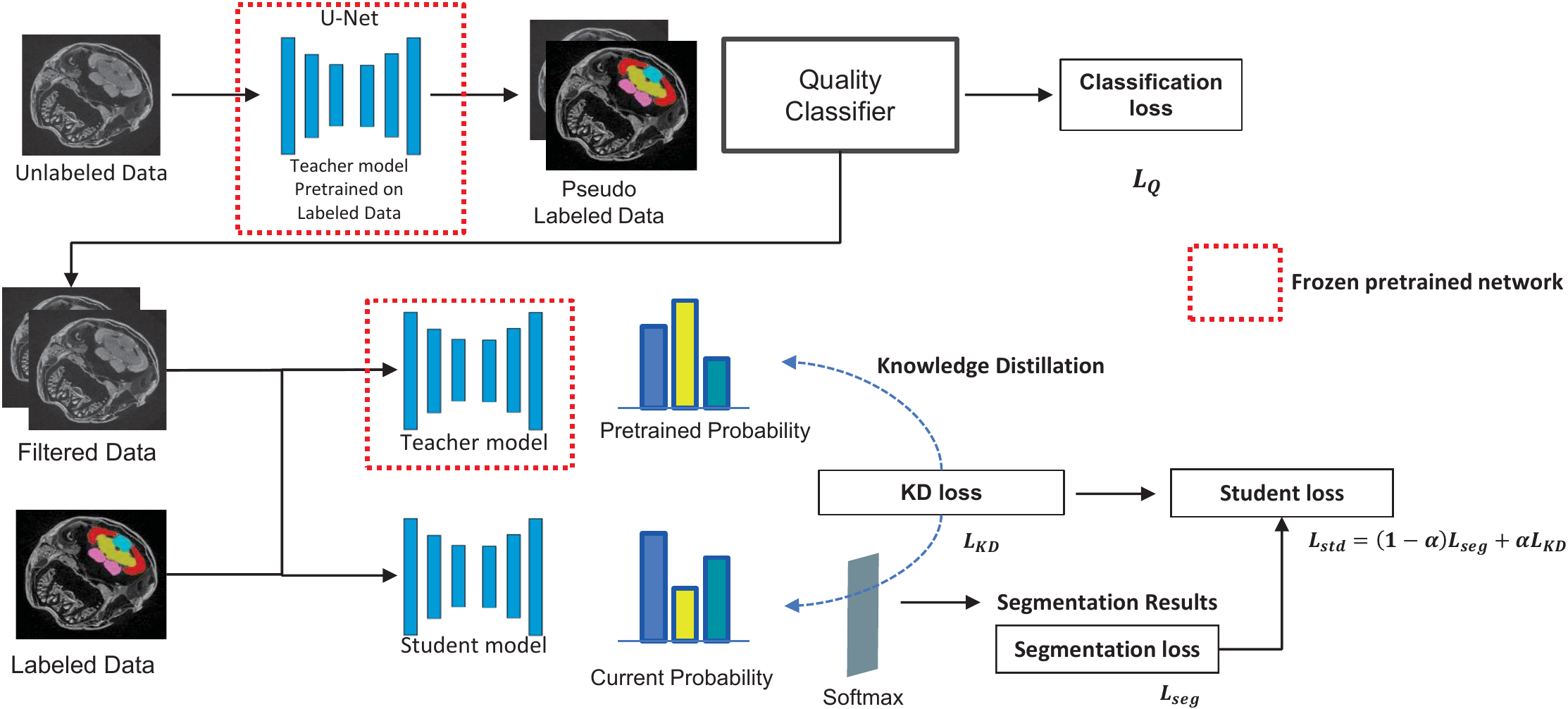}   
     \caption{Overview of the proposed framework. The teacher model is trained on the labeled data and the pseudo-labeled data are obtained by the trained teacher model. The quality classifier refines pseudo-labeled data and the student model is trained on the filtered pseudo-labeled data and labeled data by using knowledge distillation to improve the performance of the organ segmentation.}
     \label{Fig:overview_arch}
\end{figure*}

Recently, so-called semi-supervised learning has been introduced to ease the requirements for large datasets.
In \cite{shen2019cyclic}, a weakly supervised segmentation method was proposed using bounding boxes instead of segmentation masks to reduce the cost of labeling data. 
The pre-training technique aims to find a good, data-driven initialization for the model weights. This technique trains the model on large open datasets (e.g. ImageNet) before fitting it on a small task-specific dataset \cite{he2018mask}.

In contrast, Knowledge Distillation and its successor, self-training, employ the unlabeled part of the dataset \cite{hsu2019weakly}.
The core idea is to train two models, the Teacher and the Student.
The Teacher model (which could be an ensemble of models) is trained on the small labeled dataset.
Subsequently, the predictions of the Teacher model on the unlabeled part of the dataset (\emph{pseudo-labels}) are used to train the Student model.
To further improve the performance of self-training, several methods have been proposed. 
The Mean Teacher method that averages model weights to provide a better Teacher model was proposed in \cite{tarvainen2017mean}.
The NoisyStudent method proposed adding noise to the pseudo-labels to improve the generalization of the Student model \cite{xie2020self}. Zou et al. \cite{zou2018unsupervised} introduced a class-balanced self-training to select the pseudo-labels better to use for training the Student.

In this paper, we propose a knowledge distillation framework for Medaka organ segmentation in tomographic images with pseudo-label filtering. 
We utilize the pseudo-labeled data from the Teacher model pre-trained on the labeled data and adopt a Quality Classifier that learns to predict the quality of segmentation. 
The Student model is trained on the labeled data and the filtered pseudo-labels to improve the performance. 
To prevent overfitting the pseudo-labeled data and further improve the performance, we introduce a pixel-wise knowledge distillation loss that regularizes the Student. 
The proposed method is evaluated on the Medaka tomographic image dataset, and the experimental results show that our method improves the Medaka segmentation performance effectively.

The remainder of this paper is organized as follows. \cref{sec:method} provides an overview of the proposed framework and training process. In \cref{sec:results}, we present our experimental results and analysis. Finally, in \cref{sec:conclusion}, we conclude this paper.

\section{Proposed Method}
\label{sec:method}

An overview of the proposed framework is presented in \cref{Fig:overview_arch}.
It consists of three models: (1) the Teacher model, trained on the labeled data, (2) the Quality Classifier trained to distinguish good and bad pseudo-labels, and (3) the Student model trained on the filtered pseudo-labeled data along with the labeled data.
In the following subsections, we describe the components of this framework in detail.

\subsection{Models}

First, we use a U-Net \cite{ronneberger2015u} for the \textbf{Teacher model} and train it on the labeled data along with their corresponding segmentations provided by the biologists. The U-Net architecture introduces skip concatenation between the encoder and the decoder layers and provides good performance in image segmentation tasks. Let $X_L=\{x_L^i\}^{N_L}_{i=1}$ be the labeled input images with pixel-wise annotated labels $Y_L=\{y_L^i\}^{N_L}_{i=1}$. In this work, we adopted ResNet-18 \cite{he2016deep} as the encoder to achieve high performance for image classification. Then, we generate pseudo-labeled data by putting all the unlabeled data into the Teacher model. Here, we denote the unlabeled images $X_U=\{x_U^i\}^{N_U}_{i=1}$ and the teacher model $F_T: \mathcal{X_U} \rightarrow \mathcal{Y_U}$. Then, we define the pseudo labels as $Y_U=F_T(X_U)=\{y_U^i\}^{N_U}_{i=1}$. 


Among these pseudo labels, there are many slices that are harmful to be used for training the Student model. 
In this work, we adopt a \textbf{Quality Classifier} to distinguish between good and bad pseudo labels. 
To train the quality classifier, a dataset of approximately 1000 slices was manually labeled into two categories: good and bad. 
For training the Quality Classifier, each slice from the image was concatenated with the corresponding mask, where each sub-organ in the mask was represented using a separate channel.
Therefore, the refined input images are expressed as $X_R=\{x_R^i\}^{N_R}_{i=1}$ with the corresponding refined pseudo labels $Y_R=\{y_R^i\}^{N_R}_{i=1}$.

To ensure that the \textbf{Student model} performs well, we use the same architecture and model size as the Teacher.
As depicted in \cref{Fig:overview_arch}, the Student is trained on both the labeled data and filtered data, obtained by passing the pseudo-labeled through the quality classifier. This helps prevent performance degradation caused by inaccurate pseudo-labels. 
However, as the number of refined pseudo-labeled data is much higher than the number of labeled data, the Student may become biased towards the pseudo-labeled data. 
To address this, we concatenate the refined pseudo-labeled data and the labeled data $X_C=\{x_c^i\}^{N_L+N_R}_{i=1}=X_L\cup X_R$ and their corresponding labels as $Y_C=\{y_c^i\}^{N_L+N_R}_{i=1}=Y_L\cup Y_R$.
We denote the Student model $F_S: \mathcal{X_C} \rightarrow \mathcal{Y_C}$.
Then, we can express the predictive segmentation maps from the Teacher and the Student model as $\tilde{Y}_C^T=F_T(X_C)$ and $\tilde{Y}_C^S=F_S(X_C)$, respectively. 

\subsection{Training}

We define the segmentation loss on the combination of the pseudo-labeled and supervised data as follows.

    \begin{equation}
        \label{eq:segmentationloss}
        \begin{split}
            L_{seg} &= L_{CE}(F_S(X_C), Y_C),\\
            \text{where }& L_{CE}= -\sum_{k} q(k)\log(p(k)),
        \end{split}
    \end{equation}

We train the Student model using this loss to transfer knowledge from the Teacher model.
We refer to this type of training as \textbf{Pseudo-Labeling} henceforth.

To improve convergence, we adopt the idea of \textbf{Self-Training} \cite{hsu2019weakly} by initializing the Student model with the last checkpoint of the Teacher model.

Furthermore, we incorporate the concept of the \textbf{Knowledge Distillation} \cite{hinton2015distilling, yuan2020revisiting}, also known as dark knowledge distillation.
To prevent overfitting to the pseudo-labeled data and retain the knowledge from the labeled data in the pretrained Teacher model, this idea proposes directly distilling the softened labels produced by the Teacher model. 
As proposed by Hinton et al. \cite{hinton2015distilling} and Yuan et al. \cite{yuan2020revisiting}, we use temperature scaling to soften the predictions of the Teacher model.
\begin{equation}
    \label{eq:temperature}
        p^t_k(x^i;\tau) = softmax(z^t_k(x^i;\tau)) = \frac{\exp{(z^t_k(x^i)/\tau)}}{\sum_j^{K} \exp{(z^t_j(x^i)/\tau)}}
\end{equation}
where $p^t_k(x^i;\tau)$ is the $k$-th output of $i$-th pixel, $K$ is the number of segmentation classes, $z^t_k$ is the pixel-wise output segmentation logits of the pretrained teacher model and $\tau$ is the temperature to soften the predictive segmentation probability. 
Using the softened predictions, we regularize the student model by using the Kullback-Leibler (KL) divergence between the softened predictions of the Student and the Teacher for each pixel pair at the same spatial position.
The resulting knowledge distillation (KD) loss is given as follows.
    \begin{equation}
        \label{eq:KDloss}
        L_{KD}=\frac{1}{N}\sum_{i\in N} KL(p^s(x^i;\tau)\parallel p^t(x^i;\tau))
    \end{equation}
    where $N=W \times H$ is the number of pixels of the image data, $KL(\dot)$ is the KL divergence function, and $p^s(x^i;\tau)$, $p^t(x^i;\tau)$ are the output probability of the $i$-th pixel in the segmentation map from the student and the pretrained teacher models respectively.
    
The proposed method combines all the aforementioned parts together: we use the pseudo-labels further filtered by the Quality Classifier, initialize the Student with the last checkpoint of the Teacher model, and regularize the Student model via the KD loss.
The final Student loss is defined as the weighted sum of the KD loss and the segmentation loss between the Student's predictions and the ground truth. 
\begin{equation}
    \label{eq:studentloss}
    L_{std}=(1-\alpha)L_{seg} + \alpha L_{KD}
\end{equation}
where $\alpha$ controls the relative importance of different losses. By minimizing this loss, we aim to transfer the knowledge from the Teacher model to the Student model, while preventing overfitting to the pseudo-labeled data. 

\section{Experimental Results}
\label{sec:results}

\begin{figure*}[!t]
\centering
\begin{subfigure}{0.24\linewidth}
  \centering
  \includegraphics[width=\linewidth]{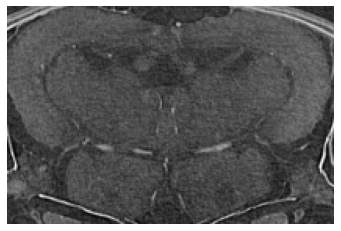}
  \caption{}
  \label{fig:gt}
\end{subfigure}%
\hfill
\begin{subfigure}{0.24\linewidth}
  \centering
  \includegraphics[width=\linewidth]{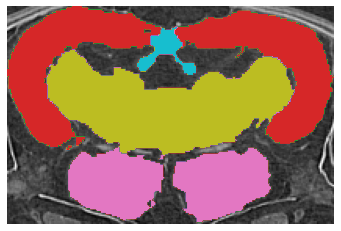}
  \caption{}
  \label{fig:teacher}
\end{subfigure}%
\hfill
\begin{subfigure}{0.24\linewidth}
  \centering
  \includegraphics[width=\linewidth]{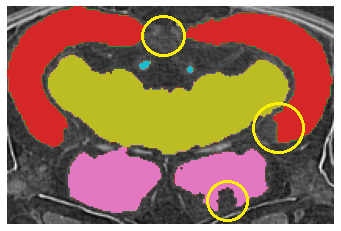}
  \caption{}
  \label{fig:stud}
\end{subfigure}%
\hfill
\begin{subfigure}{0.24\linewidth}
  \centering
  \includegraphics[width=\linewidth]{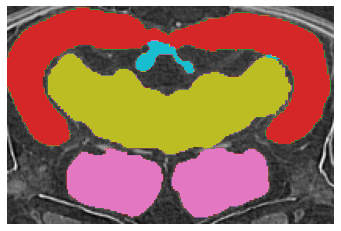}
  \caption{}
  \label{fig:kd}
\end{subfigure}
\caption{\label{fig:examples} Examples of (a) Input image, (b) ground truth, and segmentation results by (c) the Teacher (Fully Supervised), (d) the Student model by the proposed method. }
\vspace{-3mm}
\end{figure*}

In this section, we address a brief introduction to the tomographic Medaka image data and implementation details. In addition, we present the comparison results and ablation study of the proposed method in terms of the mean intersection over union (mIoU) and the dice score.

\subsection{Datasets}
The Medaka CT-scans were collected over several beam times following the protocol proposed in \cite{Weinhardt2018}.
The original volumes had the dimensions of $6000 \times 2000 \times 2000$ and were rescaled to $3000 \times 1000 \times 1000$.
The labeled dataset used to train the Teacher model consisted of 30 samples, and the unlabeled dataset consisted of a total of 582 scans from three different experiments: 169, 232, and 181 samples.
The labeled data were split into the train (75\%) and validation (25\%) sets, hence 23 volumes were used for training, and 7 were used for validation.

\subsection{Implementation Details}
We implemented the proposed method in Python using the PyTorch library. We used two NVIDIA RTX 8000 and two RTX 2080 for training and testing. Adam optimizer was used to train the proposed network with a learning rate of 3e-4. The hyperparameters in \cref{eq:studentloss} were $\alpha=0.1$ and $\tau=4$. The models were trained with a crop size of 256 and a batch size of 64, and they converged within 15 epochs. The convergence was even faster, taking only 5 epochs when initialized with the Teacher's weights.

\subsection{Baseline Comparison}



\begin{table}[!t]
    \caption{Comparison of the segmentation results between the Fully Supervised model as a baseline, and the proposed method.}
    \label{tab:labeledratio}
    \centering
    \resizebox{\columnwidth}{!}{
    \begin{tabular}{|c|cc|cc|}
        \hline
         Number of & \multicolumn{2}{c|}{mIoU (\%)}   &   \multicolumn{2}{c|}{Dice (\%)}\\
         \cline{2-5}
         labeled volumes &  proposed     & baseline &  proposed  & baseline\\ 
         \hline
         2               &  $72.5$  &  $43.2$   & $77.9$ & $48.5$ \\
         7               &  $75.6$  &  $54.7$   & $80.8$ & $55.9$ \\
         12              &  $79.8$  &  $60.1$   & $86.9$ & $65.2$ \\
         23              &  $82.4$  &  $74.8$   & $89.4$ & $82.2$ \\
         \hline
    \end{tabular}}
    \vspace{-2mm}
\end{table}

First, we present a quantitative comparison of the results of the proposed method with the Fully Supervised model as a baseline in \cref{tab:labeledratio}.
The proposed method clearly outperforms the baseline in both the low and full data regimes.
These results demonstrate that our method consistently outperformed the baseline. In addition, the proposed method using only seven labeled volumes provided better performance than the baseline using all available training data volumes, confirming the proposed method can improve the Medaka segmentation performance. Remarkably, our method using only two volumes of the labeled data achieved better results than the Fully Supervised method with 12 labeled volumes.
Second, we visually compare the results of the segmentation as shown in \cref{fig:examples}.
We marked the problematic areas of the sample with yellow circles in \cref{fig:examples} (c). 
While the prediction of the proposed method clearly has its own peculiarities of segmentation, the provided result is smoother spatially and closer to the ground truth.

\subsection{Ablation Study}

We conducted an ablation study to evaluate the impact of different components of the proposed method on the segmentation performance. The results are presented in \cref{tab:ablation}.
The proposed method performed slightly better than any of the individual components of the pipeline.
We found that the filtering of the pseudo-labels provided the most significant improvement, while the Quality Classifier training could benefit from more data.

Interestingly, the results demonstrate that the improvements of the proposed method are equal to or greater than the sum of improvements provided by the separate components.
This could mean, that different components improve the quality of the result in different ways.
This is beneficial for total improvement since improvements of the separate components do not interfere with others.

\begin{table}[!t]
\caption{Evaluation results of ablation study for the proposed framework with knowledge distillation and quality classifier using 23 labeled volumes for training.}
\label{tab:ablation}
\centering
\begin{tabular}{|l|cc|}
\hline
Method                                  & mIoU (\%) & Dice (\%) \\ \hline
Fully supervised       & $74.8$       & $82.2$  \\ 
Pseudo-Labeling       & $76.5$       & $86.0$ \\ 
\hline
Knowledge Distillation & $76.7(+0.2)$ & $86.6(+0.6)$ \\ 
Teacher Checkpoint     & $77.5(+1.0)$ & $86.8(+0.8)$ \\ 
Quality Classifier     & $81.2(+4.7)$ & $87.4(+1.4)$ \\
\textbf{Proposed}      & $82.4(+5.9)$ & $89.4(+3.4)$ \\ 
\hline
\end{tabular}
\vspace{-2mm}
\end{table}

\section{Conclusion}
\label{sec:conclusion}

In this paper, we presented a knowledge distillation framework that effectively improved the segmentation quality of Medaka fish organs through pseudo-label refinement. 
The proposed method demonstrated superior performance compared to the fully supervised training method, with a 5.9\% improvement in mIoU and a 3.4\% improvement in Dice, measured in the full data regime.
Moreover, even with only two volumes of training data, the method yielded an impressive improvement of 29.3\% mIoU, which is on par with supervised training on 12 volumes, highlighting the potential of our approach in reducing the burden of hand-drawn segmentation.
We also introduced our Quality Classifier view on the selection of pseudo-labels, which is a core component of our method and provides a substantial contribution to the overall improvement (+4.7\% mIoU out of +5.9\% in total).
In future work, we aim to extend our method by collecting more tomographic data on Medaka and other types of fish with more organisms to evaluate and improve our proposed framework. We believe that our work can provide a better understanding of future-oriented genetics.

\newpage
\section{Compliance with Ethical Standards}
The data used in this paper was collected for other projects. 
Our team has not conducted any experiments with biological samples.
As soon as the data will be published, we will update our publication to link the datasets.

\section{ACKNOWLEDGMENT}
This work was supported by the Carl-Zeiss Stiftung under the Sustainable Embedded AI project (P2021-02-009) and funded by the European Commission Project SustainML under grant agreements number 101070408. 
We acknowledge the support by the projects CODE-VITA (BMBF; 05K2016) and HIGH-LIFE (BMBF; 05K2019).
We gratefully acknowledge the data storage service SDS@hd supported by the Ministry of Science, Research and the Arts Baden-Württemberg (MWK) and the German Research Foundation (DFG) through grant INST 35/1314-1 FUGG and INST 35/1503-1 FUGG.
We acknowledge the KIT for provision of instruments at the Karlsruhe Research Accelerator (KARA) and thank the personnel of imaging beamlines.
We thank Sabine Bremer for the provided samples, and Tinatini Tavhelidse-Suck from Centre for Organismal Studies Heidelberg for the provided markup.

\bibliographystyle{IEEEbib}
\bibliography{strings,refs,zharov}

\end{document}